\documentclass[11pt]{article}

\usepackage{acl}
\usepackage{times}
\usepackage{latexsym}
\usepackage[T1]{fontenc}
\usepackage[utf8]{inputenc}
\usepackage{microtype}
\usepackage{inconsolata}
\usepackage{graphicx}
\usepackage{amsmath}
\usepackage{booktabs}
\usepackage{array}
\usepackage{xcolor}
\usepackage{multirow}

\newcommand{\khalf}{K_{1/2}}

\title{Separating Semantic Competition from Context Length in RAG Reading}

\author{
\begin{tabular}{c}
\textbf{Vyzantinos Repantis \quad Ameya Gawde \quad Harshvardhan Singh \quad Rohit Alekar} \\
\textbf{Cien Zhang \quad Svetlana Karslioglu \quad Akash Vishwakarma} \\\\
Meta Platforms
\end{tabular}
}

\begin{document}
\maketitle

\begin{abstract}
Retrieval-augmented generation (RAG) systems can respond incorrectly even when the correct passage was retrieved. The model must still read the retrieved passages and identify which one contains the answer among others that look relevant. This passage-reading model is called the \textit{reader}. Does it fail simply because the context is longer or because the other passages genuinely compete with the correct one? We introduce and demonstrate a matched-control protocol for RAG reading: we keep the number and length of passages fixed, but replace hard competitors with less competitive real passages. We apply this control across two compact open models on SQuAD. This replacement partially restores performance, with the strongest effects on F1 and answer inclusion. For Phi-2, this recovers +6.0 EM points, +7.0 answer-inclusion points, and +0.057 F1. For Qwen2.5-1.5B, it recovers +4.5 EM points, +9.0 answer-inclusion points, and +0.068 F1. To track how performance changes as competitors accumulate, we also report retention curves and summarize them with a right-censored half-life when the curves do not cross half-retention. Together, these results show the protocol isolates a competition effect distinct from context length, though the effect is clearer for F1 and answer inclusion than for exact match, and also varies with snippet length.
\end{abstract}

\section{Introduction}

Retrieval-augmented generation (RAG) systems are often evaluated by asking whether the answer appears somewhere in the retrieved context. That is necessary, but it is not sufficient: even when the right passage is present, the model still has to find and use it. Much work on long contexts treats the challenge as one of size: whether the answer fits in the context window, and whether the model can locate it as the input grows. But size is not the only thing that changes when more passages are retrieved. The passages themselves may compete: several can look relevant, and the model has to choose among them.

The component that reads the retrieved passages and produces an answer is the reader, following the standard retriever--reader decomposition in open-domain question answering \citep{chen-etal-2017-reading}. In many retrieved contexts, the reader sees several plausible passages: near-matching entities, clauses, or explanations. The correct passage is present, but it has to win a competition.

This paper isolates that competition. We ask a simple question: if we hold context length fixed, does replacing top-ranked competing passages with lower-ranked passages from the same corpus help the reader recover the answer? If yes, then the degradation is not just a length effect. It reflects semantic competition among retrieved passages.

We introduce a controlled protocol for RAG reading. Each context contains one gold snippet (a fixed-length excerpt from the passage that contains the answer) and a fixed number of distractor snippets (excerpts from other passages filtered to exclude the answer string). We compare two equal-size contexts. In the \textit{hard} condition, the distractors are top-ranked negatives retrieved by BM25, making them strong competitors for the question. In the \textit{far-control} condition, we keep the same number of passages and nearly the same input length, but replace most hard negatives with far-rank real passages from the same corpus, also filtered to exclude the answer string. This avoids a weakness of artificial filler controls: the distractors remain fluent, natural passages, but are less competitive with the question.

The same control can be instantiated with other retrievers by defining ``hard'' and ``far'' passages according to that retriever's ranking, although our matched-control experiments use BM25.

We demonstrate the protocol on two compact open readers, Phi-2 and Qwen2.5-1.5B-Instruct. Replacing hard negatives with far-rank passages partially restores F1 and answer inclusion for both models, with weaker and noisier gains on exact match. Because the hard and far-control contexts have the same passage count and nearly the same length, this recovery indicates a competition effect that is detectable when length is controlled.

\paragraph{Contributions.}
First, we introduce a matched-control protocol that separates semantic competition from context length by holding passage count and length fixed while varying how strongly the distractors compete. Second, we demonstrate it on two compact open readers, finding partial recovery concentrated in F1 and answer inclusion. Third, we report retention curves over competitor count, summarized by a right-censored half-life, to track how reader performance changes as competition grows.

\section{Protocol}

\paragraph{Task construction.}
We use extractive question answering as a controlled setting: the answer is a span of text in one of the passages. Each example consists of a question, a gold passage that contains the answer, and distractor passages drawn from the same corpus. We sample from SQuAD~1.1 \citep{rajpurkar-etal-2016-squad}. The reader sees a numbered list of passages and is instructed to answer using only those passages.

\paragraph{Hard-negative contexts.}
Let \(H\) be the number of hard negatives, i.e., top-ranked distractor snippets retrieved for the question. We vary \(H\) to trace how performance changes as the number of hard negatives increases, with each context holding one gold snippet and \(H\) hard-negative snippets. Distractors are selected from the same corpus while excluding the gold passage. The gold snippet is answer-centered when possible to keep the answer inside it. Distractor snippets are truncated to a fixed word budget. Gold position is randomized except in the gold-only baseline.

\paragraph{Matched far-rank control.}
Increasing \(H\) changes both semantic competition and total context. To isolate competition, we construct a matched \(N=20\) control. The hard condition contains the gold snippet plus 19 top-ranked negatives retrieved by BM25. The far-control condition contains the same gold snippet, four such BM25 negatives, and 15 far-rank real SQuAD snippets. All non-gold snippets in both conditions are filtered to exclude the normalized answer string. Thus, the two conditions have the same number of snippets and nearly the same input length, but differ in how strongly the distractors compete with the question. If degradation is caused only by passage count or context length, replacing hard negatives with far-rank snippets should not help. If semantic competition contributes, the far-control condition should partially recover.

\paragraph{Metrics.}
We report exact match (EM), token F1, and answer inclusion. EM requires an exact normalized match, F1 gives partial credit for token overlap, and answer inclusion checks whether any normalized gold answer appears anywhere in the reader output. Together, these metrics distinguish exact-answer recovery from partial or longer-form answer recovery.

\paragraph{Retention curves and censored half-life.}
We also summarize hard-negative sweeps with retention curves. For metric \(M\), let \(M(H)\) be performance with \(H\) hard negatives and \(M(0)\) the gold-only baseline. Retention is
\begin{equation}
R_M(H) = \frac{M(H)}{M(0)}.
\end{equation}
If the curve crosses \(R_M(H)\le 0.5\), we report the first such \(H\) as \(\khalf\). If it does not cross within the tested range, we report a right-censored value, such as \(\khalf>79\). This half-life is descriptive: it depends on the metric, retriever, and context construction, not the model alone.

\section{Experimental Setup}

\paragraph{Readers.}
We evaluate Qwen2.5-1.5B-Instruct and Phi-2. We choose compact open readers as a simple, reproducible testbed for isolating reader behavior, rather than as a claim that these models represent all RAG systems. This choice also keeps the focus on a common conflation in RAG evaluation: longer contexts and stronger competition often vary together. Our protocol separates them by varying competitor pressure directly while holding passage count and snippet length fixed.

\paragraph{Retrievers.}
The main matched-control protocol uses BM25. Hard negatives are top-ranked snippets retrieved by BM25 after excluding the gold passage. All non-gold snippets are filtered to remove snippets containing the normalized answer string. Far-rank snippets are lower-ranked real snippets from the same SQuAD corpus, filtered in the same way. We also run a dense-retriever sanity check for Qwen retention curves using BGE-small, but the matched far-rank control is BM25-based.

\paragraph{Snippet lengths.}
For the matched-control comparison across readers, we use 50-word snippets for both Qwen2.5-1.5B-Instruct and Phi-2 so that Phi-2 remains within its 2K context limit. For the Qwen retention sweep, we also evaluate 80-word snippets over \(N\in\{1,20,40,80\}\), corresponding to \(H\in\{0,19,39,79\}\). The 80-word setting provides a sensitivity check on how snippet granularity changes the measured competition effect (Section~\ref{subsec:snippet-length}).

\paragraph{Uncertainty.}
For matched hard-to-far comparisons, we report paired bootstrap confidence intervals over question IDs and sign-flip randomization \(p\)-values. EM and answer inclusion are reported in percentage points and F1 on the \([0,1]\) scale.

\section{Results}

\subsection{Retention degrades gradually}

\begin{figure}[t]
\centering
\includegraphics[width=\linewidth]{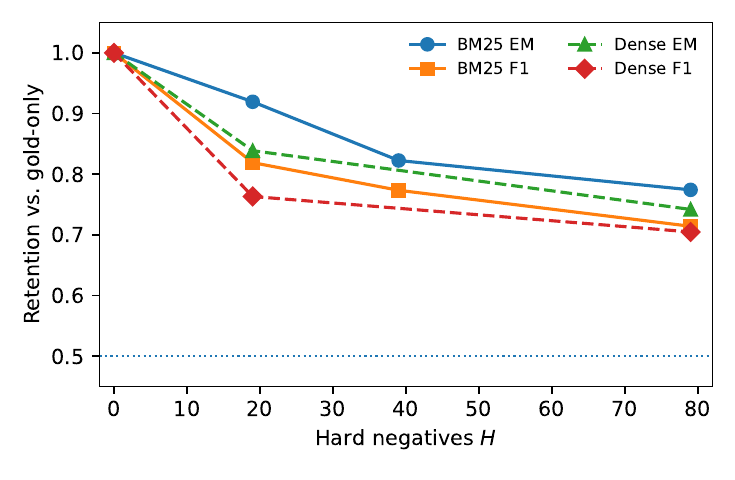}
\caption{Qwen2.5-1.5B-Instruct retention under increasing hard-negative count. Retention decreases under both BM25 and dense retrieval, but does not cross half-retention through \(H=79\).}
\label{fig:retention}
\end{figure}

Figure~\ref{fig:retention} shows Qwen2.5-1.5B-Instruct retention as hard-negative count increases. Under BM25, Qwen degrades monotonically from the gold-only condition through \(H=79\). EM falls from 31.0 at \(H=0\) to 24.0 at \(H=79\), while F1 falls from 0.575 to 0.411. The pattern holds under dense retrieval: EM falls from 31.0 to 23.0, and F1 from 0.575 to 0.405. In both retriever settings retention remains above one half through the largest tested hard-negative count, so both EM and F1 half-lives are right-censored: \(\khalf>79\). Censoring is informative here: the curves degrade steadily but never cross half-retention, so the half-life is right-censored rather than a manufactured crossing point.

\subsection{Recovery at fixed length}

Table~\ref{tab:matched-control} reports the matched snippet-50 control. Hard negatives reduce performance for both readers. For Phi-2, EM falls from 52.5 in the gold-only condition to 24.5 with 19 answer-free hard negatives. F1 falls from 0.704 to 0.459. Replacing 15 of the 19 hard negatives with far-rank real passages recovers performance to 30.5 EM and 0.516 F1. Qwen shows the same qualitative pattern. EM falls from 51.5 to 22.0 under hard negatives and F1 falls from 0.683 to 0.387. The far-control condition recovers to 26.5 EM and 0.455 F1.

\begin{table}[t]
\centering
\small
\setlength{\tabcolsep}{3.5pt}
\begin{tabular}{llrrrr}
\toprule
Reader & Condition & EM & Incl. & F1 & Tok. \\
\midrule
\multirow{3}{*}{Phi-2}
  & Gold   & 52.5 & 85.5 & 0.704 & 124 \\
  & Hard   & 24.5 & 70.0 & 0.459 & 1487 \\
  & Far    & 30.5 & 77.0 & 0.516 & 1496 \\
\midrule
\multirow{3}{*}{Qwen2.5}
  & Gold   & 51.5 & 80.5 & 0.683 & 117 \\
  & Hard   & 22.0 & 58.0 & 0.387 & 1519 \\
  & Far    & 26.5 & 67.0 & 0.455 & 1532 \\
\bottomrule
\end{tabular}
\caption{Matched snippet-50 far-rank control. The hard condition uses one gold snippet plus 19 top-ranked negatives retrieved by BM25. The far-control condition keeps \(N=20\) but replaces 15 hard negatives with far-rank real SQuAD snippets. All non-gold snippets are filtered to exclude the normalized answer string. ``Incl.'' is answer inclusion; ``Tok.'' is average input tokens.}
\label{tab:matched-control}
\end{table}

The two conditions share passage count, snippet length, and answer-string filtering, with closely matched input lengths. The recovery, therefore, cannot be explained by fewer passages, a shorter context, or answer leakage. Instead, it points to semantic competition as a distinct contributor to reader degradation.

\subsection{Clearest recovery in F1 and inclusion}

Table~\ref{tab:ci} reports paired hard-to-far deltas with 95\% bootstrap intervals and paired sign-flip \(p\)-values. For Phi-2, replacing hard negatives with far-rank snippets recovers \(+6.0\) EM points, \(+7.0\) answer-inclusion points, and \(+0.057\) F1. All three confidence intervals exclude zero. For Qwen, EM recovery is \(+4.5\) points with a confidence interval that slightly crosses zero. The clearer gains are in answer inclusion and F1: \(+9.0\) answer-inclusion points and \(+0.068\) F1, both with confidence intervals excluding zero. This metric pattern is expected: EM is strict, while F1 and answer inclusion also reward outputs that contain the answer in a longer or partially correct response.

\begin{table}[t]
\centering
\small
\setlength{\tabcolsep}{4pt}
\begin{tabular}{llrrr}
\toprule
Reader & Metric & $\Delta$ & 95\% CI & $p$ \\
\midrule
\multirow{3}{*}{Phi-2}
  & EM    & +6.0   & [1.0, 11.0]    & .036 \\
  & Incl. & +7.0   & [1.0, 13.0]    & .027 \\
  & F1    & +0.057 & [0.016, 0.099] & .009 \\
\midrule
\multirow{3}{*}{Qwen2.5}
  & EM    & +4.5   & [-0.5, 9.5]    & .136 \\
  & Incl. & +9.0   & [2.5, 15.5]    & .010 \\
  & F1    & +0.068 & [0.015, 0.122] & .012 \\
\bottomrule
\end{tabular}
\caption{Paired hard-to-far recovery under the matched snippet-50 control. EM and answer inclusion are percentage-point deltas. F1 is a raw-score delta.}
\label{tab:ci}
\end{table}

\subsection{Effect varies with snippet length}
\label{subsec:snippet-length}

The gold-only baseline differs between the 80-word retention sweep and the 50-word matched-control experiments. This is expected: the 50-word gold snippets are answer-centered and remove more surrounding context, making the gold-only condition easier. We, thus, compare hard and far-control conditions only within the same snippet-length protocol. The snippet-80 far-control uses the same answer-string filtering as the matched-control runs. Its hard condition, therefore, differs slightly from the unfiltered retention sweep: gold-only and hard-19 EM are 31.0 and 29.5, respectively.

This distinction also explains why the measured recovery differs across Qwen settings. At 80-word snippets, Qwen shows much weaker hard-to-far recovery: EM recovery shrinks from +4.5 points at snippet-50 to +0.0 points at snippet-80, while F1 recovery shrinks from +0.068 to +0.020. Longer snippets can preserve more answer-supporting context, reducing ambiguity among candidate passages and leaving less room for far-control recovery. The measured recovery depends on snippet length and context construction, and should be reported with them rather than read as a model-only property.

\section{Related Work}

Long-context benchmarks such as Lost in the Middle and RULER test whether models can access information in long inputs \citep{liu2024lost,hsieh2024ruler}. They expose position, length, and task-complexity effects, but they do not hold length fixed while replacing top-ranked retrieved competitors with less competitive real passages.

RAG evaluation frameworks often separate retrieval-side and generation-side quality, such as whether retrieved contexts are relevant and whether the generated answer is faithful or correct \citep{es2023ragas}. RAG engineering analyses also identify failures where evidence is missing, not retrieved highly enough, not included in context, or present but not extracted by the model \citep{barnett2024seven}. Our protocol isolates one mechanism behind such failures: the gold evidence is present, but semantically competing passages can prevent the reader from selecting it. By holding length and passage count fixed, we test this competition directly.

NoLiMa argues that long-context evaluation should go beyond literal matching: when question and needle share little lexical overlap, models must infer latent associations, and performance drops as context grows \citep{modarressi2025nolima}. Our control studies a different axis: holding length fixed while changing how strongly retrieved distractors compete with the question.

\section{Limitations}

These experiments use one extractive QA dataset and two compact open readers, so they are a controlled demonstration of the protocol rather than a claim about all RAG systems. Other corpora, retrievers, and frontier-scale readers may show different competition profiles. The strongest recovery signals are F1 and answer inclusion. EM recovery is significant for Phi-2 but weaker for Qwen, so the effect is best read as partial restoration of answer selection and span quality, not of exact-match accuracy. Further, our matched far-rank control is BM25-based. The dense check in this paper is limited to the retention curve. Finally, censored half-life is only a compact description of a retention curve. When curves do not cross half-retention, the right report is the full curve plus a censored value, not an invented crossing point.

\section{Conclusion}

RAG reader evaluation should measure more than whether the answer is present somewhere in context. We introduced a matched-control protocol for separating semantic competition from context length: compare hard retrieved negatives with far-rank real passages while holding passage count and snippet length fixed. On SQuAD-based experiments with two compact open readers, hard negatives substantially reduce performance, and replacing most of them with far-rank passages partially restores F1 and answer inclusion. Semantic competition is a distinct contributor to reader degradation beyond length alone, with magnitude depending on metric and snippet length.

\bibliography{custom}

\end{document}